\documentclass[letterpaper, 10 pt, conference]{ieeeconf}  

\IEEEoverridecommandlockouts                              

\usepackage{graphics, graphicx} 
\usepackage{amsmath} 
\usepackage{amssymb}  
\usepackage{tabularx, array, makecell}
\usepackage{multirow}
\usepackage{hyperref}
\usepackage{subcaption}
\usepackage{color}
\usepackage[T1]{fontenc}
\usepackage{pdfpages}

\title{\LARGE \bf
Guiding Collision-Free Humanoid Multi-Contact Locomotion using Convex Kinematic Relaxations and Dynamic Optimization
}

\author{Carlos Gonzalez$^{1}$ and Luis Sentis$^{1}$
\thanks{$^{1}$Authors are with the Department of Aerospace Engineering and Engineering
Mechanics, The University of Texas at Austin, TX 78712, USA. 
Emails: \{carlos.gonzalez, lsentis\}@utexas.edu.}%
}

\begin{document}
\null%
\includepdf[pages=-]{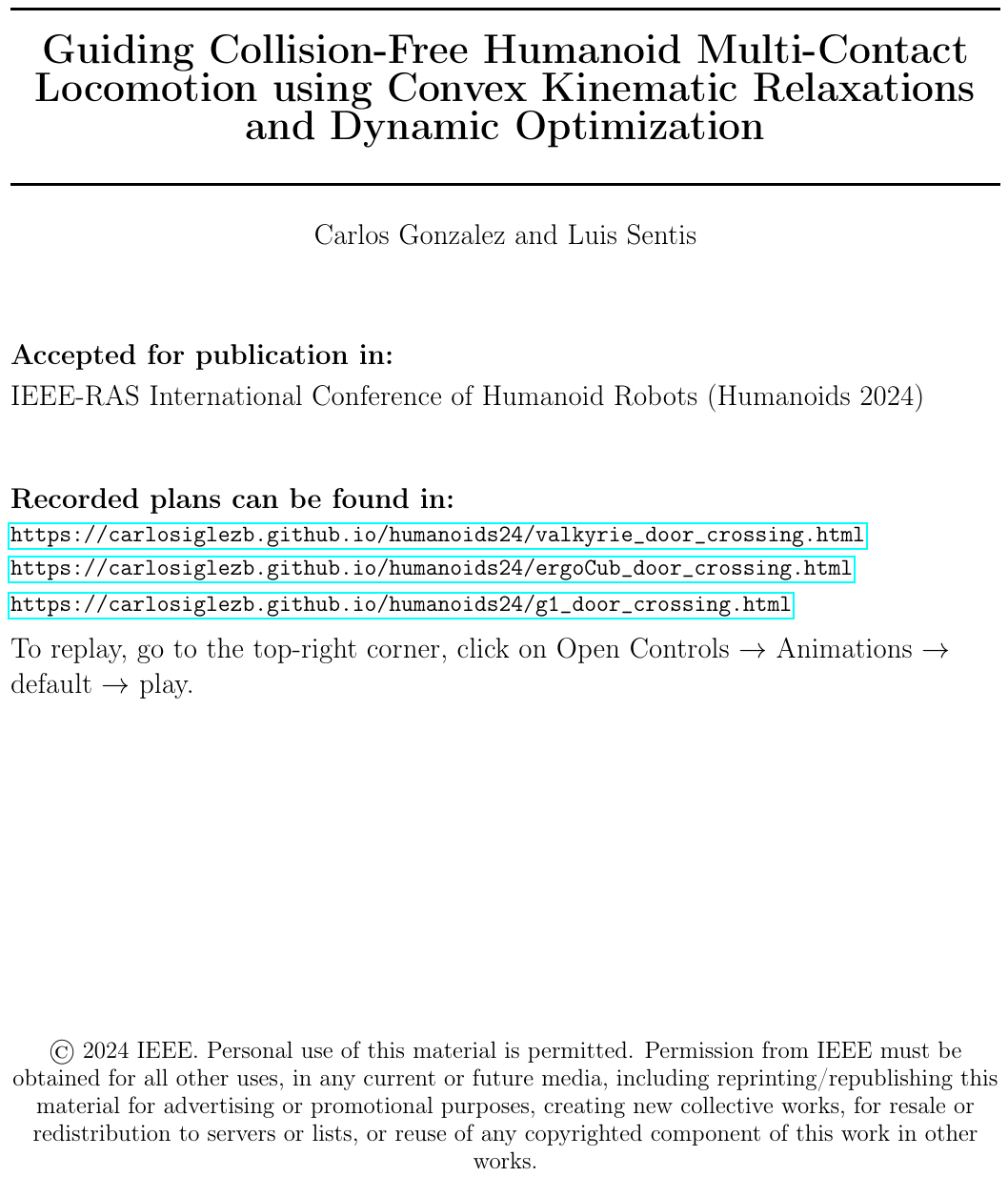}

\maketitle
\thispagestyle{empty}
\pagestyle{empty}

\begin{abstract}
Humanoid robots rely on multi-contact planners to navigate a diverse set of environments, 
including those that are unstructured and highly constrained. 
To synthesize stable multi-contact plans within a reasonable time frame, most planners 
assume statically stable motions or rely on reduced order models. 
However, these approaches can also render the problem infeasible in the presence
of large obstacles or when operating near kinematic and dynamic limits.
To that end, we propose a new multi-contact framework that leverages recent advancements in relaxing 
collision-free path planning into a convex optimization problem, extending it to be applicable to humanoid multi-contact navigation. 
Our approach generates near-feasible trajectories used as guides in
a dynamic trajectory optimizer, altogether addressing the aforementioned limitations.
We evaluate our computational approach showcasing three different-sized humanoid robots traversing
a high-raised naval knee-knocker door using our proposed framework in simulation. Our approach can generate motion plans within a few seconds consisting of several multi-contact states, including dynamic feasibility in joint space.
\end{abstract}

\section{INTRODUCTION}

\begin{figure}
    \centering
    \includegraphics[width=0.75\linewidth]{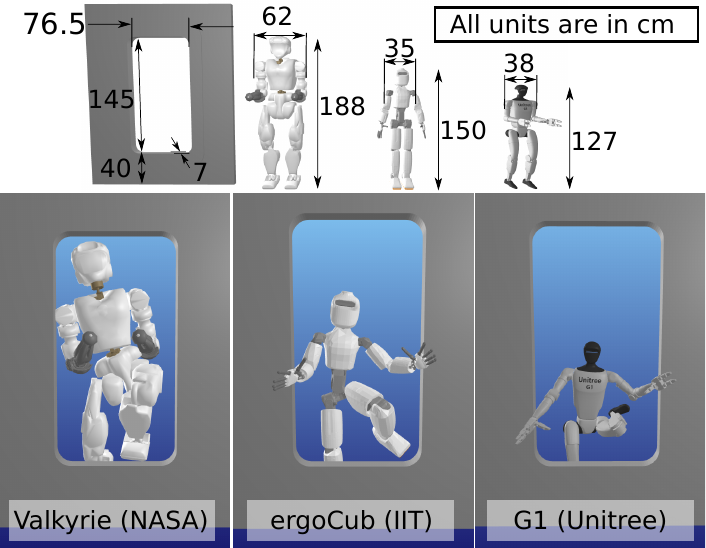}
    \caption{Example of a constrained environment requiring dexterous whole-body motions while operating
    near kinematic limits to traverse it. 
    This highlights both the complexity of 
    of a big humanoid to avoid all obstacles and of a smaller one to pass the high step of the door.
    }
    \label{fig:knee-knocker-versatile}
\end{figure}

Until recently, planning collision-free motions using optimization-based methods
had been predominantly approached as a nonlinear program
(NLP) due to the non-convex nature of collision-avoidance
constraints~\cite{Dai2014,Haffemayer2024ModelArm,Marcucci2023MotionOptimization}. 
Recent advancements in convex optimization have introduced tight
relaxations that make the collision-avoidance problem  
convex, leading to smooth and efficiently computed motions~\cite{Marcucci2023MotionOptimization}.
While these methods have matured in applications such as 
exploring contacts during manipulation~\cite{Graesdal2024TowardsManipulation}, their
potential in legged multi-contact locomotion --- where 
highly articulated systems have to navigate constrained environments
with obstructive obstacles,
as shown in Fig.~\ref{fig:knee-knocker-versatile} ---
has not been yet explored.


Classic humanoid multi-contact locomotion approaches often decompose the planning problem
into two or more phases. The first phase involves planning a contact 
sequence towards the goal. 
The remaining phase(s) use the planned contacts to construct kinematically and dynamically feasible motions between contact states. The final output is the desired feasible 
torques~\cite{Wensing2023Optimization-BasedRobots}.
In this work, we assume a given contact sequence that does not necessarily satisfy
static equilibrium conditions. This approach allows our planner to search over a broader range
of motions, including both dynamically and statically stable motions. 
We then proceed with two separate stages. First, by building on
the work of~\cite{marcucci23_fpp}, we use feasibility proxies to search for 
reachable and collision-free motions of robot key frames using 
a convex formulation. 
We then use the resulting trajectories as 
references (or guides) to facilitate the planning of whole body motions in the collision-free space. The contributions of our work are described in the detail in the next section. 


The paper is organized as follows: Section~\ref{sec:related_work} 
compares our approach with respect to prior work in detail, 
Section~\ref{sec:preliminaries} summarizes the fast path planning algorithm that we exploit~\cite{marcucci23_fpp}, 
Section~\ref{sec:approach} presents our proposed framework
and the details of each submodule therein, 
Section~\ref{sec:results} presents our quantitative and qualitative simulation results,
and we conclude with a discussion and future work in 
Section~\ref{sec:future_work}.

\section{RELATED WORK}\label{sec:related_work}

Strategies in legged multi-contact locomotion have significantly evolved over the last few decades. Common dichotomies during this evolution include
the use of sampling-based versus optimization-based methods, 
the metrics to characterize physical balance, and
the application of model-based versus learning-based approaches. 

Sampling-based planners have historically been attractive due to their 
ease of implementation and straightforward collision detection. 
Hence, multi-contact planners have adopted these methods to varying degrees. 
For instance, in the aforementioned ``stance-before-motion'' approach,  
sampling-based approaches are often utilized during the contact planning phase,
This includes planning the collision-free root path~\cite{Tonneau2018AnRobots,Kumagai2019BipedalFeasibility} or determining
footsteps~\cite{Kumagai2018EfficientMotion,Lin2020RobustPrediction}. 
Sometimes sampling-based 
methods~\cite{Ferrari2023Multi-contactFramework,Kingston2023ScalingSearch} also assist in-between-contact motions.
On the other hand, optimization-based methods are appealing for their ability to generate
smooth motions while taking into account kinematic and dynamic 
constraints. For instance, optimization-based methods are commonly used in inverse kinematics solvers
to resolve constrained motions between pre-specified contact configurations~\cite{Tonneau2018}.
They can also be used to circumvent the ``stance-before-motion'' approach
by jointly searching for both contacts and motions through solving an
Optimal Control Problem (OCP)~\cite{Wang2020Multi-FidelityLocomotion,Winkler2018GaitParameterization,Graesdal2024TowardsManipulation,marcucci23_fpp,Wang2022LearningPlanning}. 
Hybrid approaches that combine sampling- and optimization-based methods 
have also been explored to efficiently generate multi-contact manipulation and 
locomotion, although have been mostly assessed in 
quadrupeds~\cite{Sleiman2023VersatileLoco-Manipulation,Jelavic2023LSTP:Systems}.
Another optimization-based approach involves solving
Mixed-Integer Programs, which solve for both discrete and continuous variables. Although these are often
computationally expensive, recent tight relaxations have made
them more tractable and shown promise to blend discrete and continuous planning~\cite{gcs}. 
However, aside from footstep planning, these methods have not yet been fully extended to legged locomotion.
Our approach builds on some of the optimization-based methods proposed in~\cite{marcucci23_fpp} and its follow-up version~\cite{gcs}. 

Planning motions with balance constraints offers a trade-off between 
safely reaching the goal and overcoming
challenging paths, such as those requiring motions with momentum. 
While static equilibrium assumptions have vanished from walking strategies, 
many multi-contact locomotion algorithms still lean towards planning statically stable 
motions~\cite{Ferrari2023Multi-contactFramework,McCrory2023GeneratingVisualization,Tonneau2018AnRobots,Kumagai2019BipedalFeasibility}. 
Other strategies relax this constraint by ensuring that the robot's Center of Mass (CoM) 
is within kinematic reach 
with respect to the active contacts~\cite{Wang2024OnlineApproximation,Wang2020Multi-FidelityLocomotion,Wang2022LearningPlanning},
while others focus on reachability approaches using sequenced reduced order 
models~\cite{Zhao2022ReactiveEnvironments}.
In our work, we first generate motions such that the robot's torso
remains within kinematic reach of its stance foot. We then verify
that the robot stays up and that no constraints are violated during the dynamic 
motion planning process.

The rise of learning methods has brought new potential for legged locomotion.
Due to the extensive literature on this topic, readers are referred 
to~\cite{Ha2024Learning-basedPerspectives} for a recent survey on learning-based methods.
An similarly informative survey on optimization-based methods 
is provided by~\cite{Wensing2023Optimization-BasedRobots}.
While learning-based methods have rapidly evolved, predicting their scalability outside of trained environments, especially in more constrained settings, remains challenging.

\subsection{Contribution}
Our contributions go as follows:
\begin{enumerate}
    \item Devising the first collision-free convex kinematic path planning algorithm for humanoid multicontact planning. We achieved this by leveraging \cite{marcucci23_fpp} and exploiting the use of multiple interconnected planning particles to find robot body paths in Cartesian space.   
    \item Introducing a streamlined approach for providing the aforementioned collision-free trajectories to a full-body multi-contact planner throughout the planned trajectory. We achieve this by employing the Differential Dynamic Programming (DDP) method in~\cite{Mastalli2020Crocoddyl:Control} applied to the full and long horizon trajectories.
    
\end{enumerate}
In addition, the framework is designed to produce results quickly with respect to the planning horizon. 

\section{PRELIMINARIES: FAST PATH PLANNING}\label{sec:preliminaries}

Our approach to compute the approximate kinematic collision-free trajectories 
builds upon the fast path planning algorithm presented in~\cite{marcucci23_fpp}, 
which we summarize here. 

Given a set of $K$ axis-aligned and collision-free boxes in a $d$-dimensional space, 
$\mathcal{B}_k:=\{x \in \mathbb{R}^d ~| ~l_k \leq x \leq u_k \}$ for
$k=1, \cdots, K$, a distance-weighted graph is constructed by
connecting the intersection of neighboring boxes. The lengths of the edges are minimized, and the resulting graph is used to search for the shortest path between an initial and final point in the collision-free space. 
The index of the $j^{\mathrm{th}}$ box in the sequence of this path is $s_j$.
The interior nodes along the corresponding path of $N$ segments, represented by
$\{z_j\}_{j=1}^{N-1}$, where $z_j \in \mathbb{R}^d$, are then adjusted
to further shorten the path 
by solving the following convex optimization problem:
\begin{equation}
\begin{aligned}
     \underset{z}{\mathrm{minimize}}      & \quad \sum\limits_{j=1}^N \| z_j - z_{j-1} \|_2  \\ 
     \mathrm{subject~to}    & \quad z_{j-1}, z_j \in \mathcal{B}_{s_j}, \quad j = 1, \cdots, N \\
                            & \quad z_0 = p^{\mathrm{init}}, \quad z_N = p^{\mathrm{term}}
\end{aligned}
\label{eq:polygonal}
\end{equation}
Next, a new sequence of boxes is found via another OCP to further reduce
the path length. By iterating over these two problems, an optimized path and box
sequence is obtained. Finally, the optimal box sequence is used to find a smooth collision-free path parameterized 
by B\'ezier curves by solving a Quadratic Program (QP) with the following characteristics:
\begin{equation}
    \begin{aligned}
        \mathrm{minimize}   & \quad \sum_{d=1}^D \mathrm{Bezier}~d~\mathrm{derivative} \\
        \mathrm{subject~to} & \quad \mathrm{boundary~conditions} \\
                            & \quad \mathrm{box~containment} \\
                            & \quad \mathrm{derivative~is~Bezier} \\
                            & \quad \mathrm{continuity} \\
                            & \quad \mathrm{continuous~derivative}
    \end{aligned}
    \label{eq:smooth}
\end{equation}
For the detailed description of the cost 
and constraints, see~\cite{marcucci23_fpp} . While~\cite{marcucci23_fpp} also includes an additional re-timing step to 
further improve transition times within a fixed path, we omit this step 
as it is not employed in our formulation.

\section{MAIN APPROACH}\label{sec:approach}

\subsection{Framework Overview}
%
%
\begin{figure}
    \centering
    \includegraphics[width=0.95\linewidth]{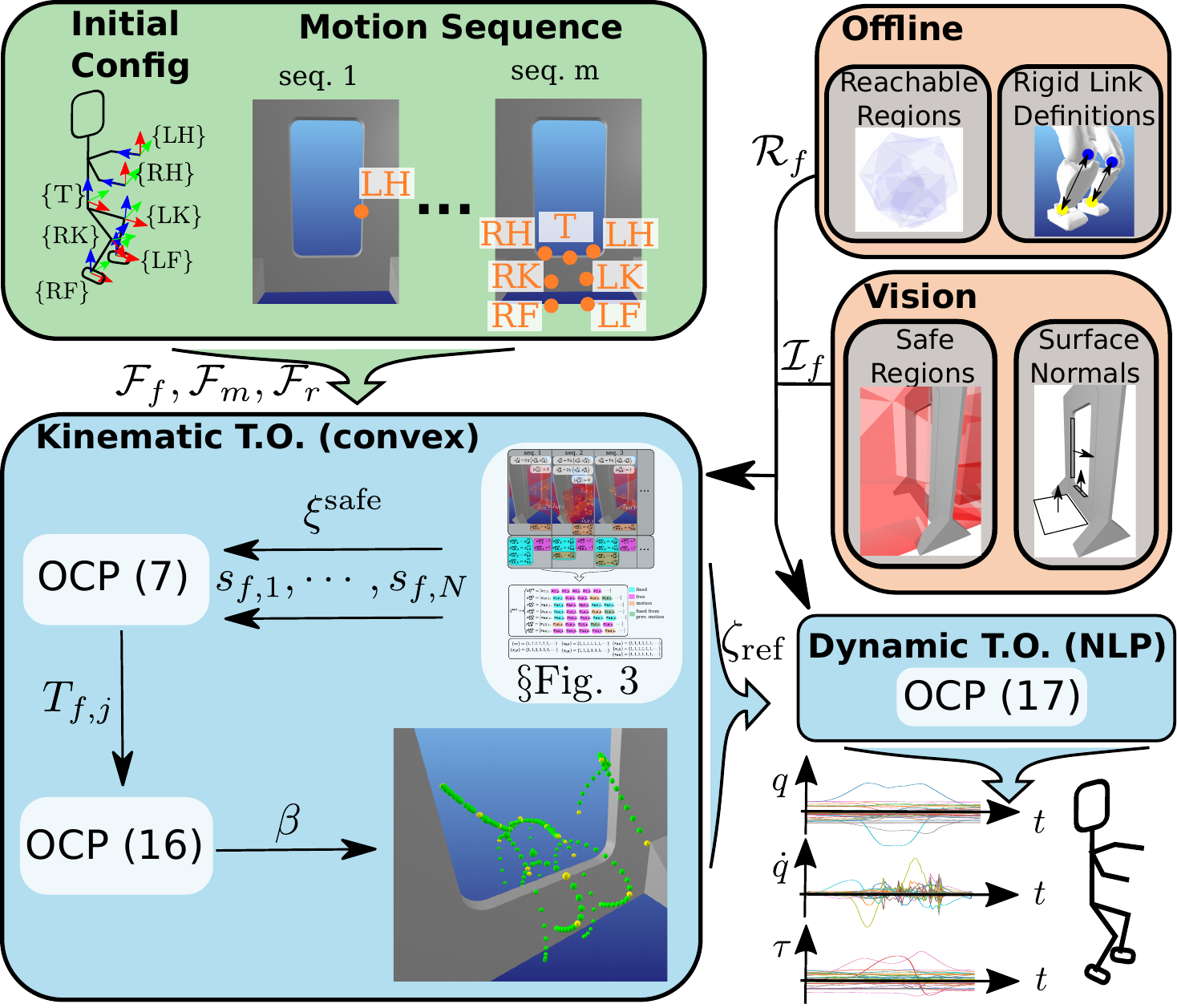}
    \caption{Block diagram showing a comprehensive overview of our proposed framework. The labels
    in \textsf{OCP} indicate the equations that describe those optimization problems.}
    \label{fig:framework}
\end{figure}
Our proposed framework
requires the user to specify in advance frames on the robot that will be
used for planning (e.g., contact points or collision-free waypoints). 
Here, we consider frames attached to the robot's torso, feet, knees, and hands, and will be referred as frames set: 
$\mathcal{F}:= \{ \mathrm{T}, ~\mathrm{LF}, ~\mathrm{RF}, 
    ~\mathrm{LK}, ~\mathrm{RK}, ~\mathrm{LH}, ~\mathrm{RH} \}$.

The overall framework is shown in Fig.~\ref{fig:framework}.
The motion generation algorithms (shown in blue) rely on geometric information
about both the robot and the environment (shown in orange). The robot's geometric information only needs to be computed once, and can therefore be done offline. This includes the kinematically 
reachable regions of each frame $\mathcal{F} \backslash \mathrm{T}$ w.r.t. the torso frame, and the
constant distance between frames attached to each rigid link. 
In this case, the feet frames are attached to the ankles\footnote{When planning foot positions, we offset the desired waypoints by the distance between the foot and the ground.}, so we compute and store the distance between the knee and feet frames.
%
The geometric information about the environment consists of 
simplified geometric representations of the obstacles, which can be obtained through tools such as~\cite{brazil_omni3d}, and their
corresponding planar regions or surface normals, where applicable.
From these representations of the geometric obstacles, we construct safe, collision-free 
regions using the method described in~\cite{Deits2015ComputingProgramming}.
The motion sequence provided by the user (shown in green) is specified by sequential segments
where frames can be set as either 
\textit{fixed} frames, $\mathcal{F}_f$, if they must not move during the corresponding segment, 
\textit{motion} frames, $\mathcal{F}_m$, indicating the desired intermediate
position, or 
as \textit{free} frames, $\mathcal{F}_r$, if neither is applicable or desired. 

With this information, a plan of approximately feasible kinematic motions is computed
and given as reference to a dynamic trajectory optimization (TO) solver. 
We next explain how the aforementioned inputs are processed and used in the
kinematic and dynamic TO formulations.


\subsection{Kinematic Trajectory Optimization}
We define kinematically feasible those trajectories that keep each
frame (including the torso) within kinematic \textit{reach} and 
within the \textit{collision-free} space at all times. 
Since we do not check for collisions of rigid bodies explicitly, 
we use a simple proxy to account for rigid links.
We represent the position of the humanoid's frames by
$x_f \in \mathbb{R}^3$ for all $f \in \mathcal{F}$. 


\subsubsection{Reachable End-Effector Constraints}
The reachable regions of the robot's end effectors and its knees are fixed w.r.t. the 
torso frame\footnote{In this work we assume
that humanoids with torso yaw keep it at a zero rotation, when generating the reachable region. This avoids rotating the reachable regions, 
which would break the convexity of the formulation.}.
Each of these regions will likely be non-convex, but
can be approximated as polyhedron using sampling and fitting heuristics
as in~\cite{Tonneau2018AnRobots}. 
Thus, the end-effectors' and knees' reachable regions
can be approximated by a region:
\begin{equation*}
    \mathcal{R}_f := \{x \in \mathbb{R}^3 ~ | ~ H_f x \leq d_f \},
\end{equation*}
where $H_f \in \mathbb{R}^{p \times 3}$ and $d_f \in \mathbb{R}^p$ describe
the $p$ half-spaces that define the polytope of frame 
$f \in \mathcal{F} \backslash \mathrm{T} $ w.r.t. the torso frame. 
As the humanoid moves, these regions translate with the robot according to the set:
\begin{equation}
    \mathcal{R}_f' := \{ x \in \mathbb{R}^3 ~|~ H_f (x - x_T) \leq d_f \}
    \label{eq:polytope_shifted}
\end{equation}
for all $f \in \mathcal{F} \backslash T$, where $x_T$ is the position of the torso frame.

\subsubsection{Reachable Torso Constraint}
The reachability of the torso is verified if it is
reachable from the stance end-effector. 
Since the position of the stance end-effector remains fixed during its corresponding sequence,
the reachable region~\eqref{eq:polytope_shifted} for that stance end-effector implicitly
constrains the torso to remain reachable. Hence, no additional constraints are needed to
satisfy reachability of the torso.
%
%
%
%

\subsubsection{Collision-Free Regions and Sequence}\label{subsec:iris-sequence}
We first generalize the formulation introduced in Sec.~\ref{sec:preliminaries}
from axis-aligned boxes to polytopes. We construct these polytopes using the
Iterative Regional Inflation by Semidefinite programming
(IRIS) algorithm~\cite{Deits2015ComputingProgramming} in Cartesian space, and 
refer to them simply as IRIS regions.

Each frame must be able to move from their initial to their final positions by
traversing an initially unknown sequence of IRIS regions. 
Creating these regions requires the selection of a seed point around which the IRIS region is grown. We used a simple approach for this: each frame, has two default seeds to grow the IRIS regions, one at the position of the initial frame and another at the approximate final location of the frame. If these two regions do not intersect (or their intersecting volume is small), an additional seed is sampled in between these two seeds near an empty free space. This process can be repeated if no safe path is found, but we also present alternatives in Sec.~\ref{sec:future_work}.

Since different frames will have to traverse a different number of IRIS regions to get to their respective goal locations, we must construct a sequence associating each frame to an IRIS region
throughout the duration of the plan. In other words, for each frame
$f \in \mathcal{F}$ we must assign a corresponding IRIS region at all times, 
which we define (with a slight abuse of notation) as:
\begin{equation}
    s_{f}(t) = \begin{cases}
            s_{f,1}(t) & \quad t \in [0, t_{f,1}] \\
            \quad \vdots &  \\
            s_{f,N_f}(t) & \quad t \in (t_{f, N_f-1}, t_{f,N_f}] \\            
            \end{cases}
            \label{case:iris_seq}
\end{equation}
%
%
Ideally, each frame would have a different number of transitions, 
$N_f \in \mathbb{N}$, hence also a different sequence 
and different transition times, $t_{f, k}~ \forall ~k = 1, \cdots, N_f-1$. 
Instead of optimizing for these sequences along with the
trajectory, we first determine a (sub-)optimal sequence order, then distribute their times
along each IRIS region, and finally optimize the trajectories based on the established sequence.

\begin{figure}
    \centering
    \includegraphics[width=0.95\linewidth]{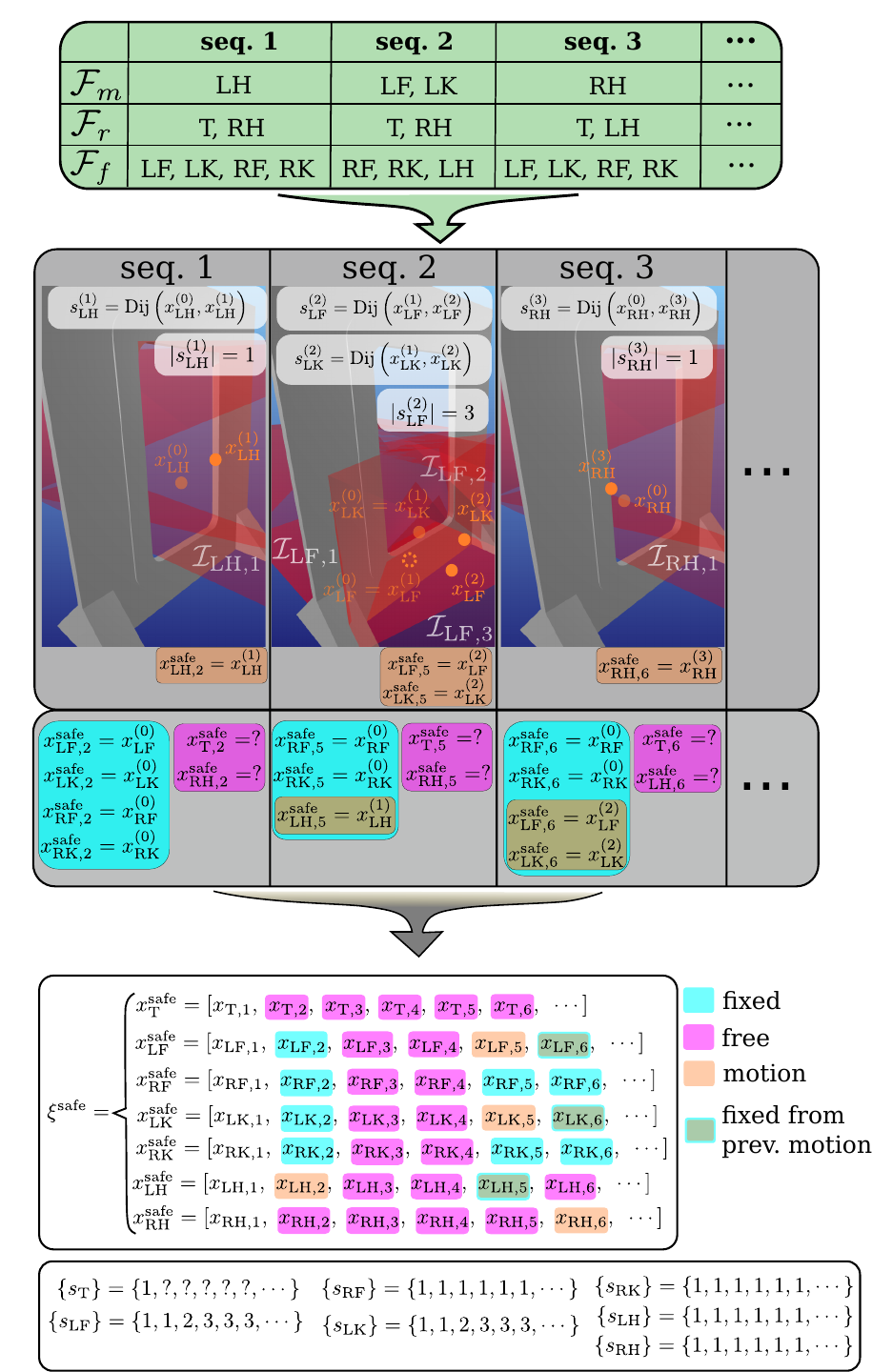}    
    \caption{Procedure followed to construct the safe waypoints, $\xi^{\mathrm{safe}}$ 
    and IRIS sequences, $\{s_{f}\} ~\forall~f \in \mathcal{F}$, from the given
    \textit{fixed}, \textit{motion}, and \textit{free} frames.}
    \label{fig:iris_sequence}
\end{figure}
%
In the absence of specified intermediate waypoints, as in~\cite{marcucci23_fpp},
the sequence order can be found through the Dijkstra's algorithm. 
To account for the intermediate waypoints specified through the motion
sequence, additional steps are required. 
Conceptually, we first discretize the trajectories of all frames 
into $P$ points each, i.e., into $x_{f,i} \in \mathbb{R}^3$ with $i = 1, \cdots, P$
for each frame $f \in \mathcal{F}$.
Then, we set the intermediate waypoints for each frame $f \in \mathcal{F}$
at their corresponding discretized time index in $x_{f}^{\mathrm{safe}}$.
This process is exemplified next.

Consider the example contact sequence shown at the top of Fig.~\ref{fig:iris_sequence}. 
We start by processing the desired motion frames in sequential order. 
For each frame, we find a collision-free path to transition from its last-known locations to
its next specified location by doing a shortest path search over the IRIS regions
(e.g., by using the Dijkstra's algorithm). In our example, during sequence 1,
the only motion frame is the left-hand.
We find that we must traverse
through one IRIS region (labeled as $\mathcal{I}_{\mathrm{LH},1}$) to transition from its initial position,
$x_{\mathrm{LH}}^{(0)}$, 
to its desired position, $x_{\mathrm{LH}}^{(1)}$. Consequently, all frames are assigned 
one variable of size 3 in sequence 1, namely, $x_{f, 2} ~\forall~f \in \mathcal{F}$. The \textit{fixed} frames are already specified, while the  \textit{free} frames are to be determined. 
We move on to sequence 2. There are two motion frames: the left foot and the knee.
The last specified location of the left-foot was its initial position, as it was
designated to be a fixed frame in sequence 1. We solve
for the shortest path from its previous location, $x_{\mathrm{LF}}^{(0)}$,
to its next desired location, $x_{\mathrm{LF}}^{(2)}$, and determine that it must
traverse three IRIS regions (labeled 
$\mathcal{I}_{\mathrm{LF}, 1}, ~\mathcal{I}_{\mathrm{LF}, 2}, ~\mathcal{I}_{\mathrm{LF}, 3}$). 
Thus, all frames are assigned three variables of size 3 each in sequence 2,
namely, $x_{f,3}, ~x_{f,4}$ and $x_{f,5} ~\forall~f \in \mathcal{F}$,
of which only the free frames are to be solved for. The indices of these 
traversed IRIS regions are stored within its sequence $\{s_{\mathrm{LF}}\}$.
The same process is applied to the left-knee frame.
Moving on to sequence 3, we have one motion frame: the right-hand. 
Its last-specified location was its initial position since it was specified as a free 
frame during sequences 1 and 2. Its shortest path from $x_{\mathrm{RH}}^{(0)}$
to $x_{\mathrm{RH}}^{(3)}$ consists of traversing one IRIS region (labeled
$\mathcal{I}_{\mathrm{RH}, 1}$). This procedure is repeated until we reach
the last sequence.
It can be seen that several positions are deduced from previously
defined fixed frames, while others still need to be solved for. Therefore, in this example,
some IRIS regions are marked as undefined (denoted by ``$?$'') since their next
desired position has not yet been specified.
Additionally, the dimension of the optimization variable is only determined  after we have reached the end of the sequence.



\subsubsection{Rigid Link Constraints}
\begin{figure}
    \centering
    \includegraphics[width=0.9\linewidth]{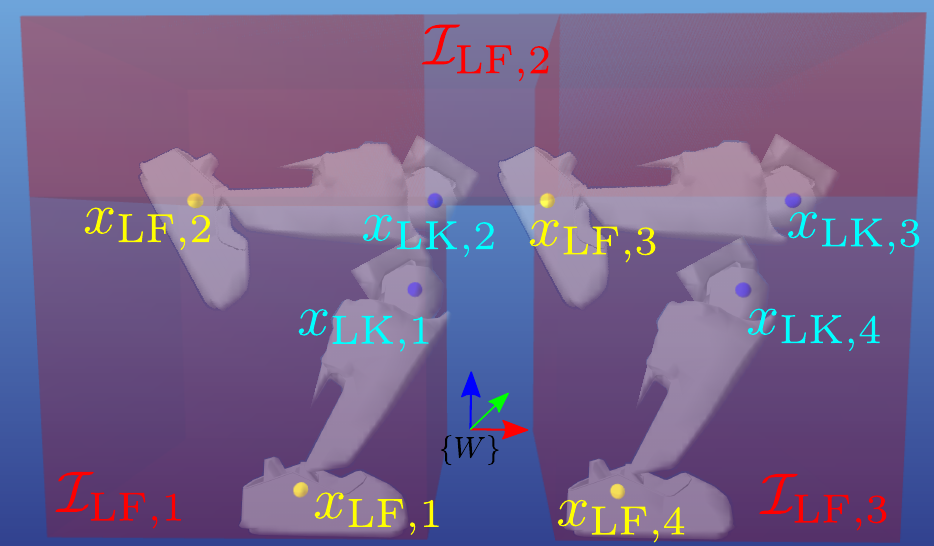}
    \caption{Side view of left foot IRIS regions, 
    $\mathcal{I}_{\mathrm{LF}, (\cdot)}$,
    constructed around a rectangular obstacle in between 
    the initial ($x_{(\cdot), 1}$) and final ($x_{(\cdot), 4}$) frame positions.
    This shows a typical outcome of including the rigid link constraint relaxation to find 
    proxy collision-free motions, 
    resulting in the distance between
    the LF and LK frames being roughly constant while remaining inside the 
    IRIS regions at the specified times. 
    }
    \label{fig:iris-rigid-relax}
\end{figure}

Our frame set $\mathcal{F}$ contains two auxiliary frames, namely the left and right
knee frames. These allow us to plan collision-free trajectories more accurately 
by constraining them to live within the safe (collision-free) regions.
However, they cannot move
independently from the feet frames as they are attached to the same rigid link.
The constraint achieving this condition for the left leg is
\begin{equation}
    \| x_{\mathrm{LF}} - x_{\mathrm{LK}} \|_2 = \ell_{\mathrm{L}}
    \label{eq:rigid_link_eq}
\end{equation}
where $\ell_{\mathrm{L}}$ denotes the expected length between the left foot and 
left knee frames. However, this is a non-convex constraint. 
To keep our OCP convex, we introduce a convex relaxation. 
First, we relax~\eqref{eq:rigid_link_eq} into an inequality constraint. 
Then, we introduce a convex penalty barrier term penalizing the knee and foot
frames for being close to each other at time instance $i$ as follows:
\begin{equation}
    J_{\mathrm{rigid}, i}^{\mathrm{L}} (x_{\mathrm{LF}, i}, x_{\mathrm{LK}, i}) := -\sum\limits_{k=1}^3 
            w_k \log \left( x_{\mathrm{LK}, i}^{(k)} - x_{\mathrm{LF}, i}^{(k)} \right)
    \label{eq:rigid_link_cost}
\end{equation}
where the superscript indicates the $k^{\mathrm{th}}$ component of the variable.
As a result, the variables associated with the knee and foot frames are designed
to be as far apart as possible without exceeding the length of the rigid link they
are attached to. A similar relaxation is done for the right leg. 
A sample set of points resulting from this relaxation while being constrained
to lie within their associated IRIS regions is shown in 
Fig.~\ref{fig:iris-rigid-relax}.

\subsubsection{Kinematic Problem Formulation}
We now define the sequence of convex optimization problems that we solve 
to find kinematically feasible trajectories.

Let us define the optimization variable at time $i$ to be
\begin{equation*}
\xi_i := \left[ x_{\mathrm{T}, i}^\top ~x_{\mathrm{LF}, i}^\top ~x_{\mathrm{RF}, i}^\top 
    ~x_{\mathrm{LK}, i}^\top ~x_{\mathrm{RK}, i}^\top ~x_{\mathrm{LH}, i}^\top 
    ~x_{\mathrm{RH}, i}^\top \right]^\top \in \mathbb{R}^{21}.
\end{equation*}
%
%
Our goal is to find the points $\xi_i ~\forall ~ i=2, \cdots, P-1$ that go from 
the initial to the final desired frame locations 
in minimum path distance while respecting the aforementioned
feasibility constraints. 
We do this by solving the following extended version of problem~\eqref{eq:polygonal} 
taking into account the reachability, collision-free, 
and rigid link constraints while respecting the 
processed (user-specified) motion sequence:
\begin{subequations}
    \begin{align}
     \underset{\xi}{\mathrm{minimize}}      
            & \quad \sum\limits_{i=2}^P \sum\limits_{f \in \mathcal{F}} \| x_{f,i} - x_{f,i-1} \|_2 
            + \sum\limits_{j \in \{\mathrm{L,R} \} } J_{\mathrm{rigid}, i}^{j} (\xi_i) \\ 
     \mathrm{subject~to}    
            & \quad x_{f, i-1}, x_{f,i} \in \mathcal{I}_{s_{f,i}} \\
            & \quad x_{f,i} \in \mathcal{R}_{f,i}' \\
            & \quad \| A_{j, i} \xi_i \| \leq \ell_j, \quad j \in \{\mathrm{L, R} \} \\
            & \quad x_{a,i} = x^{\mathrm{safe}}_{a,i}, \quad \forall ~a \in \mathcal{F} \backslash \mathcal{F}_{r,i}
    \end{align}
    \label{eq:polygonal_rigid}
\end{subequations}
with constraints applied at all $i = 2, \cdots, P$ and $f \in \mathcal{F}$.
$A_{j, i} \in \mathbb{R}^{3 \times 21}$ is the inequality constraint version 
of~\eqref{eq:rigid_link_eq} written in matrix form at time index $i$, 
$x^{\mathrm{safe}}_{a,i}$ is the safe waypoint of frame $a$ at time index $i$,
constructed in Sec.~\ref{subsec:iris-sequence}, and $\mathcal{F}_{r,i}$ is the 
\textit{free} frame set at time index $i$.

Based on the resulting paths, we distribute the transition times for 
each frame ($t_{f,k}$ in~\eqref{case:iris_seq})
according to its relative length in each IRIS region.
As a result, the transition times can change for each frame. 
This allows motions to be more versatile while remaining feasible.

Lastly, we seek to obtain a smooth parameterization of these paths for all frames. 
In order for the resulting trajectories to remain inside the collision-free space, we resort to
B\'ezier curves, which have the property that all intermediate points
of the curve remain inside the convex hull formed by the control points. 
Given the obtained IRIS sequences, the segments' durations on each IRIS region, 
and the intermediate waypoints, we formulate another convex problem to find such 
parameterized curves. Let us denote the optimization variable 
$\beta_{f,i} := \left[b_{f,i}^{(1)} ~\cdots ~b_{f,i}^{(M)}  \right]$
denoting the 
$M \in \mathbb{N}$ B\'ezier 
control points of frame $f \in \mathcal{F}$ at segment $i \in \{1, ~\cdots, ~S\}$, 
with each $b_{f,i}^{(\cdot)} \in \mathbb{R}^{3}$. 
Then, the constraints in~\eqref{eq:smooth} are adapted to each frame as explained next.
The boundary conditions constraint is applied using the safe waypoints slightly
differently in the \textit{fixed} and \textit{motion} frame sets:
\begin{align}
    \beta_{s,i} &= \mathrm{rep}(x_{s,i}^{\mathrm{safe}}, M), \quad \forall ~s \in \mathcal{F}_{f,i} 
            \label{eq:bezier-safe-fixed}\\
    b_{s,i}^{(1)} &= x_{s,i}^{\mathrm{safe}} ~ \text{and} ~  b_{s,i}^{(M)} = x_{s,i+1}^{\mathrm{safe}}
    ~\forall ~s \in \mathcal{F}_{m,i}   \label{eq:bezier-safe-motion}
\end{align}
for all $i=1, \cdots, S-1$, where the operator $\mathrm{rep}(x, M)$ repeats the 
vector $x$, $M$-times. The box containment constraint is replaced with the IRIS set containment
constraint applied to all control points:
\begin{equation}
    \beta_{f,i} \in \mathcal{I}_{s_{f,i}} \quad \forall ~f \in \mathcal{F}, ~\forall ~i =1, \cdots, S
    \label{eq:bezier-iris-containment}
\end{equation}
The last three constraints in~\eqref{eq:smooth} are properties of the B\'ezier curves
and are thus applied in the same manner but only within each frame throughout the time 
indices $i=1, \cdots, S$. The reachability
and rigid-link relaxation constraints are given, respectively, by:
\begin{align}
    \beta_{f,i} &\in \mathcal{R}_{f,i}' \quad \forall ~f \in \mathcal{F} \label{eq:bezier-reachability} \\
    \|b_{\mathrm{LF}, i}^{(1)} - b_{\mathrm{LK},i}^{(1)}  \|_2 &\leq \ell_{\mathrm{L}} 
        \label{eq:bezier-rigid-left} \\ 
    \|b_{\mathrm{RF}, i}^{(1)} - b_{\mathrm{RK},i}^{(1)}  \|_2 &\leq \ell_{\mathrm{R}} 
        \label{eq:bezier-rigid-right}   
\end{align}
Lastly, we consider the normal of the plane in contact with frame $f$ at time instance $i$ by
$\hat{n}_{f,i}$, and, where desired, approach the 
contact surfaces parallel to its normal for better contact stability. We do this by defining the constraints:
\begin{align}
    \hat{n}_{f,i} \times \dot{b}_{f,i}^{M-1} &= 0 \label{eq:bezier-zero-tangential-vel} \\
    -w_f \hat{n}_{f,i}^\top \dot{b}_{f,i}^{M-2} &\geq 0 \label{eq:bezier-slow-normal-vel}
\end{align}
where $\dot{b}$ denotes the velocity of the B\'ezier curve and $w_f \in \mathbb{R}$ 
is a velocity-based weight.
Thus, our full OCP becomes:
\begin{equation}
    \begin{aligned}
        \underset{\beta}{\mathrm{minimize}}       & \quad \sum\limits_{f \in \mathcal{F}} 
                                \sum\limits_{d=1}^{D} \alpha_d \sum\limits_{i=1}^{S} 
                                \int_{t_{f,i-1}}^{t_{f,i}} \| \beta_{f,i}^{(d)} \|_2^2  \\
        \mathrm{subject~to}     & \quad \eqref{eq:bezier-safe-fixed}, \eqref{eq:bezier-safe-motion},
                                \eqref{eq:bezier-iris-containment}, \eqref{eq:bezier-reachability},
                                \eqref{eq:bezier-rigid-left}, \eqref{eq:bezier-rigid-right},
                                \eqref{eq:bezier-zero-tangential-vel}, \eqref{eq:bezier-slow-normal-vel} \\
                                & \quad \mathrm{derivative~is~Bezier} \\
                                & \quad \mathrm{continuity} \\
                                & \quad \mathrm{continuous~derivative}
    \end{aligned}
    \label{eq:kin_bezier}
\end{equation}
%
%
with the cost 
denoting the sum of the weighted $d$-derivatives of the B\'ezier curves 
of each frame throughout all segments.


\subsection{Dynamic Trajectory Optimization}
Although the trajectories resulting from~\eqref{eq:kin_bezier} are (approximately)
kinematically feasible, they
may still be dynamically infeasible. For instance, the forces needed to follow the trajectories
might not respect the torque limits of the actuators or friction constraints. 
Hence, we use these trajectories as soft references for a full-body dynamics solver
to realize dynamically consistent whole-body motions 
around them.

\subsubsection{Contact Sequencing}

Consider the motion sequence provided by the user.
Each sequence may require a different degree of 
precision in the dynamic computation. For instance, simply moving a hand to 
reach a wall requires less discretization affinity than crossing 
a high obstacle with a leg. Thus, we consider varying time discretization
dimensions for each motion segment. 

\subsubsection{Dynamic Constraints}
The constraints we consider to determine the motion to be dynamically feasible
are listed next.
First, any two bodies in contact with each other
must respect friction. Here, we assume a static Coulomb friction model. 
Second, joints cannot move past their limits and, 
third, actuators must respect their torque limits.

\subsubsection{Dynamic Problem Formulation}
Let us denote the system state by $y = \left[ q~ \dot{q} \right]$ 
composed of the robot state $q \in \mathbb{R}^{n_q}$ and 
velocities $\dot{q} \in \mathbb{R}^{n_v}$,
and the (control) torques by $\tau \in \mathbb{R}^{n_u}$.
We find a suitable dynamically feasible trajectory by solving the following
NLP:
\begin{subequations}
    \begin{align}
        \underset{y, \tau}{\mathrm{minimize}}       & \quad \sum\limits_{j=1}^{m} \sum\limits_{i=1}^{N_j - 1} 
                                        J_i(y, \tau; \zeta_{\mathrm{ref}}) + J_{N_j} (y_{N_j}) \label{sub:dyn_cost} \\
        \mathrm{subject~to}     & \quad y_{k+1} = f(y_k, \tau_k), \quad k=1, \cdots, N_m  \label{sub:dyn_dynamics} \\
                                & \quad \lambda_k \in \mathcal{K},  \qquad \qquad ~~~ k = 1, \cdots, N_m \label{sub:dyn_contact}   \\
                                & \quad y_k \in \mathcal{Y}, \quad \tau_k \in \mathcal{U}, \quad k = 1, \cdots, N_m \label{sub:dyn_joint_lims} 
    \end{align}
    \label{eq:dynamic_formulation}
\end{subequations}
where the running costs $J_i$ in~\eqref{sub:dyn_cost}, which penalize tracking errors, 
include the smooth references
$\zeta_{\mathrm{ref}}$ obtained by constructing the Cartesian space trajectories
from the the B\'ezier curves computed in~\eqref{eq:kin_bezier}.
The dynamics~\eqref{sub:dyn_dynamics} are computed at the joint level with different 
time discretizations
over different segments for a total of $N_m = \sum\limits_{j=1}^m N_j$ time steps.
\eqref{sub:dyn_contact} is the friction constraint applied at the corresponding contacts,
and~\eqref{sub:dyn_joint_lims} are the joint and torque limits.

An advantage of parameterizing~\eqref{eq:kin_bezier} with
B\'ezier curves is that 
the reconstruction of $\zeta_{\mathrm{ref}}$ is exact
even if the discretization times for each motion segment are different.

\section{RESULTS}\label{sec:results}
%
%
%
\begin{figure*}
    \centering
    \begin{subfigure}[t]{0.25\linewidth}
        \href{https://carlosiglezb.github.io/humanoids24/valkyrie_door_crossing.html}{\includegraphics[width=\linewidth]{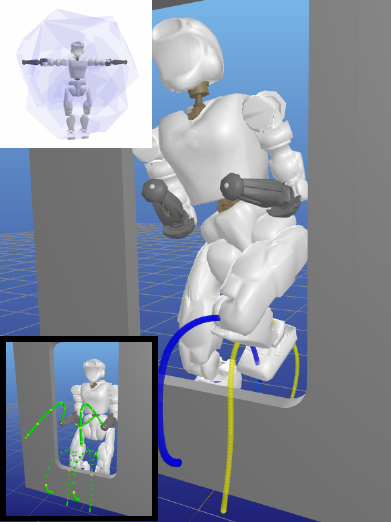}}
    \end{subfigure}
    \begin{subfigure}[t]{0.25\linewidth}
    \href{https://carlosiglezb.github.io/humanoids24/ergoCub_door_crossing.html}{\includegraphics[width=\linewidth]{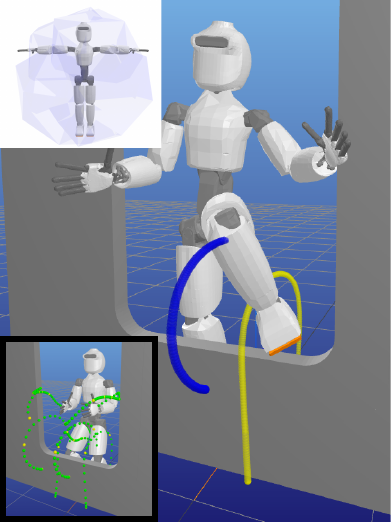}}    
    \end{subfigure}
    \begin{subfigure}[t]{0.25\linewidth}
        \href{https://carlosiglezb.github.io/humanoids24/g1_door_crossing.html}{\includegraphics[width=\linewidth]{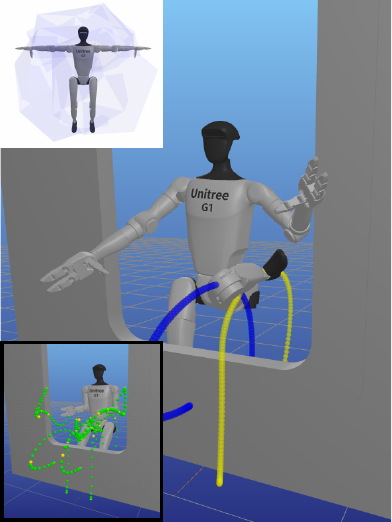}}
    \end{subfigure}
    \caption{Reachable regions, kinematically feasible trajectories for all robot frames, 
    and dynamically feasible motions
    of \href{https://carlosiglezb.github.io/humanoids24/valkyrie_door_crossing.html}{Valkyrie} (left), 
    \href{https://carlosiglezb.github.io/humanoids24/ergoCub_door_crossing.html}{ergoCub} (middle), and 
    \href{https://carlosiglezb.github.io/humanoids24/g1_door_crossing.html}{G1} (right) as they traverse 
    the knee-knocker door  using our framework.}
    \label{fig:res-kin-dyn-motions}
\end{figure*}
%
%

We consider the knee-knocker environment shown in Fig.~\ref{fig:knee-knocker-versatile} as our
constrained environment to traverse. This environment  
requires most humanoids to operate 
near their kinematic limits. 
The dimensions of the knee-knocker door are taken from a naval door 
\href{https://www.ajmarine.org.uk/index.php/4/}{seller}
with a 40 cm step height. 

The performance of our algorithm is evaluated by analyzing the computational time
of the different modules shown in Fig.~\ref{fig:framework}.
Its versatility is demonstrated by testing it on three 
different-sized humanoids: 
NASA's Valkyrie~\cite{Radford2014Valkyrie:Robot}, 
IIT's ergoCub~\cite{Shah2023Towards2.0}, and Unitree's G1.
All simulations 
are run on a desktop with an Intel Core i7-8700
equipped with 12 cores @ 3.7GHz (no parallelization was performed)\footnote{Our code is open-sourced: \texttt{https://github.com/carlosiglezb/ \\
PyPnC/tree/master/pnc/planner/multicontact}.}.

%
%
\begin{table}
    \centering
    \begin{tabular}{|r|c|c|c|}
        \multicolumn{4}{c}{} \\
        \hline
         &      & Motion frames    & Free frames \\
         \hline
        \parbox[c]{10mm}{\multirow{5}{*}{\makecell{ergoCub \\ $\&$ \\ G1}}} 
        & seq 1 & LH, T & RH  \\
        & seq 2 & LF, LK      & T, RH \\
        & seq 3 & RH, T         & RH \\
        & seq 4 & T, RF, RK    & LH \\
        & seq 5 & RH, LH        & None \\
        \hline
        \parbox[c]{10mm}{\multirow{3}{*}{\makecell{Vakyrie}}} 
        & seq 1 & LF, LK, LH, RH        & T  \\
        & seq 2 & T, RF, RK, LH, RH     & None \\
        & seq 3 & all                   & None \\ 
        \hline
    \end{tabular}
    \caption{Contact sequence provided to the different humanoids to pass through the knee-knocker door.}
    \label{tab:contact-seq}
\end{table}
%
%

For all humanoids, we construct each frame's reachable region by sampling 10$^4$ 
poses within the robot's kinematic limits and set the number of polytope planes
in~\eqref{eq:polytope_shifted} to $p=30$. 
Due to the different humanoid dimensions
with respect to the tight space in the knee-knocker, we used two separate contact
sequences, as shown in Table~\ref{tab:contact-seq}. These contacts were initially chosen
heuristically to achieve higher stability by choosing handholds at the height of the
robot's shoulder, where its reachable region intersected the door.
Each sequence motion was set
to last $3$ seconds, which are used to compute the B\'ezier parameters, and the objective in~\eqref{eq:kin_bezier} was set to minimize jerks.
The optimization parameters for each humanoid robot type were tuned separately, although one set of gains often served as a good initial
set of parameters for the other robot types. 
We used CVXPY to solve these sets of problems with the CLARABEL solver and, 
when it failed to find a solution, we used SCS with lower accuracy tolerances 
as a back-up solver.

The main tuning parameters in~\eqref{eq:dynamic_formulation} were the weights
of the tasks and the numerical discretization time steps. 
We used Croccodyl's FDDP solver to solve~\eqref{eq:dynamic_formulation} 
with the inequality constraints implemented as soft constraints.
The provided initial guess was a static pose at the initial segment location with 
the required torques to achieve static balance.
The cost weights were tuned following standard rules, i.e., 
gauging the trade-offs between the competing costs (tasks). 
The discretization time steps were chosen based on the complexity of each motion segment, 
e.g., 
for G1: $\{N_j\}_{j=1}^5 = \{80, 150, 80, 150, 100\}$. 

The resulting reachable polytopes, the kinematically feasible trajectories, and
the corresponding dynamically feasible motions are shown in Fig.~\ref{fig:res-kin-dyn-motions}.
It can be seen that the polytopes provide a fair approximation of the reachable space
of the robot w.r.t. its torso. 
The kinematic trajectories are displayed in the bottom-left corners as green spheres every 
0.36 seconds, with
the yellow spheres indicating the initial and ending positions 
when transitioning from one IRIS region to the next.
It can be seen that the trajectories are smooth, collision-free, and 
also slow down as they approach a new contact surface, providing smooth contact transitions. 
The expanded figures show a snapshot of each humanoid as they cross the knee-knocker with their left 
foot. Their respective LK and LF frame reference trajectories are shown in blue and yellow, respectively. 
It can be seen that 
these trajectories serve as guides while the robots diverge from them as needed.
As a result, each humanoid type chooses to exploit their own kinematic structure. 
For example, the G1 robot is shown folding its leg towards its body to follow the reference 
trajectories without compromising other 
tasks\footnote{Recorded animations of these motions can be found at \texttt{https://carlosiglezb.github.io/humanoids24/ \\ \{valkyrie, ergoCub, g1\}\_door\_crossing.html} and played by clicking on 
Open Controls $\rightarrow$ Animations $\rightarrow$ default $\rightarrow$ play.}.

%
%
%
%
\begin{table}
    \centering
    \begin{tabular}{|c|c|c|c|c|c|}
        \multicolumn{6}{c}{} \\
        \hline
        \multirow{2}{*}{Humanoid} & \multirow{2}{*}{IRIS} & \multicolumn{2}{c|}{Kin. Feas.} & Dyn. Feas. & Plan \\
        \cline{3-4}
            &  & OCP~\eqref{eq:polygonal_rigid}  & OCP~\eqref{eq:kin_bezier}   &  OCP~\eqref{eq:dynamic_formulation} & duration \\
        \hline
        Valkyrie    &   0.111     & 0.009      & 0.362       & 3.773 & 9\\
        \hline
        ergoCub     &   0.123     & 0.001      & 0.151       & 64.004 & 15 \\    
        \hline
        G1          &   0.117     & 0.001      & 0.186       & 19.798 & 15\\    
        \hline
    \end{tabular}
    \caption{Compute times (all in seconds) of the different planning components for each robot depicted in Fig.~\ref{fig:res-kin-dyn-motions} with contact sequences from Tab.~\ref{tab:contact-seq}.}
    \label{tab:compute-times}
\end{table}
%
%

The computation times of the different components of our proposed framework are shown 
in Table~\ref{tab:compute-times}. Solving~\eqref{eq:polygonal_rigid}
is the fastest step of the pipeline, while the dynamic TO is the
most time consuming. We note that the NLP solution is computed the fastest in Valkyrie,
primarily because its motion sequence consists of only three motion sequences
with the fewest \textit{free} frames. In contrast, the other two humanoid robots have five sequences with a higher number of \textit{free} frames.
Hence, the NLP solution is the fastest for G1 relative to its complexity.
We suspect this is due to its relatively stronger joints and wider range of motion. 
In practice, the proposed end-to-end planning process would be executed every time a challenging terrain requiring multi-contact is encountered. 
The computed joint
positions, velocities, and torque can then be used as references for a lower
level controller (e.g., a full-body MPC) to track in realtime.

\subsection{Limitations}
The proposed framework provides an intuitive, visual, and fast algorithm to plan
full-body dynamically feasible multi-contact motions guided by collision-free paths. 
One advantage is that due to the convex relaxations, the kinematic collision-free
paths do not need careful initialization. 
However, our approach has some limitations:
(i) the convex approximation of the reachable regions~\eqref{eq:polytope_shifted} contain 
some self-collisions, mistakenly considering them as feasible,
(ii) poorly choosing the seeds to generate the IRIS regions can impact the resulting
trajectories from the kinematic TO, hence impacting the dynamic TO solution, 
(iii) we do not specify balance constraints in our kinematic TO, thus we cannot guarantee
that these guide references will promote stable motions, and
(iv) for motions operating near the robot's limits, the large number of constraints 
in the OCPs can
sometimes render the problems infeasible; 
hence, when exploring new motion sequences, some constraints had to be disabled 
to produce feasible solutions.
We propose methods to address each of these in our future work discussion.

\section{DISCUSSION AND FUTURE WORK}\label{sec:future_work}
We presented a framework to plan humanoid locomotion around collision-free regions 
in multi-contact scenarios.
Our approach follows the ``stance-before-motion'' structure while leveraging optimization
tools and assuming a given contact sequence. 
We showcased our results in a challenging constrained
environment requiring careful planning around kinematic and dynamic limits where the usage
of hand contacts is crucial to the successful traversing of the environment.
Instead of interpolating between configurations, solving a single NLP, or using a reduced order model, 
we decompose the problem into a (convex) kinematic TO followed by a full-body dynamics TO problem.
In the kinematic TO problem, we quickly search for reachable, collision-free trajectories. 
We then use these to guide the full-body dynamic TO through the reachable, collision-free space. 
The proposed framework easily transfers to
different humanoids and also outputs visual queues helpful to the user for design 
and testing alternative contact sequences.


This work presents several interesting venues for future work. Foremost,
we seek to address the above-mentioned limitations of our 
framework. 
Self-collisions 
can be avoided by 
promoting motions in areas of higher manipulability~\cite{Yoshikawa1985ManipulabilityMechanisms.}
or with a higher occupancy measure (similar to what is done in~\cite{Carpentier2018MulticontactRobots}
with centroidal occupancy measures)
when solving the kinematic OCP~\eqref{eq:kin_bezier}, which is currently not time-intensive.
The location of the IRIS seeds can be optimized by using dedicated algorithms, e.g.,~\cite{WernerICRA24Cliques}.
In order to address stability guarantees, one can consider
the kinematic CoM constraint polytope approximation~\cite{Tonneau2018},
which can be (almost readily) included in our formulation as a proxy.

Other interesting future efforts include learning
multi-contact foot- and
hand-holds to robustly traverse constrained and potentially moving environments.
We also plan to include impact dynamics and add a low-level controller prior to proceeding
with experiments on a real robot.

\addtolength{\textheight}{-1cm}   





\section*{ACKNOWLEDGMENT}
This work was supported by the Office of Naval Research (ONR), Award No. 
N00014-22-1-2204.


\bibliographystyle{IEEEtran}
\bibliography{references}

\begin{thebibliography}{10}
\providecommand{\url}[1]{#1}
\csname url@samestyle\endcsname
\providecommand{\newblock}{\relax}
\providecommand{\bibinfo}[2]{#2}
\providecommand{\BIBentrySTDinterwordspacing}{\spaceskip=0pt\relax}
\providecommand{\BIBentryALTinterwordstretchfactor}{4}
\providecommand{\BIBentryALTinterwordspacing}{\spaceskip=\fontdimen2\font plus
\BIBentryALTinterwordstretchfactor\fontdimen3\font minus \fontdimen4\font\relax}
\providecommand{\BIBforeignlanguage}[2]{{%
\expandafter\ifx\csname l@#1\endcsname\relax
\typeout{** WARNING: IEEEtran.bst: No hyphenation pattern has been}%
\typeout{** loaded for the language `#1'. Using the pattern for}%
\typeout{** the default language instead.}%
\else
\language=\csname l@#1\endcsname
\fi
#2}}
\providecommand{\BIBdecl}{\relax}
\BIBdecl

\bibitem{Dai2014}
H.~Dai, A.~Valenzuela, and R.~Tedrake, ``{Whole-body Motion Planning with Simple Dynamics and Full Kinematics},'' in \emph{International Conference on Humanoid Robots}, 2014, pp. 295--302.

\bibitem{Haffemayer2024ModelArm}
A.~Haffemayer, A.~Jordana, M.~Fourmy, K.~Wojciechowski, F.~Lamiraux, and N.~Mansard, ``{Model predictive control under hard collision avoidance constraint for a robotic arm},'' in \emph{Ubiquitous Robots}, 2024.

\bibitem{Marcucci2023MotionOptimization}
T.~Marcucci, M.~Petersen, D.~von Wrangel, and R.~Tedrake, ``{Motion planning around obstacles with convex optimization},'' \emph{Science Robotics}, vol.~8, no.~84, p. eadf7843, 2023.

\bibitem{Graesdal2024TowardsManipulation}
\BIBentryALTinterwordspacing
B.~P. Graesdal, S.~Y.~C. Chia, T.~Marcucci, S.~Morozov, A.~Amice, P.~A. Parrilo, and R.~Tedrake, ``{Towards Tight Convex Relaxations for Contact-Rich Manipulation},'' 2024. [Online]. Available: \url{http://arxiv.org/abs/2402.10312}
\BIBentrySTDinterwordspacing

\bibitem{Wensing2023Optimization-BasedRobots}
P.~M. Wensing, M.~Posa, Y.~Hu, A.~Escande, N.~Mansard, and A.~D. Prete, ``{Optimization-Based Control for Dynamic Legged Robots},'' \emph{IEEE Transactions on Robotics}, vol.~40, pp. 43--63, 2023.

\bibitem{marcucci23_fpp}
\BIBentryALTinterwordspacing
T.~Marcucci, P.~Nobel, R.~Tedrake, and S.~Boyd, ``{Fast Path Planning Through Large Collections of Safe Boxes},'' 2023. [Online]. Available: \url{http://arxiv.org/abs/2305.01072}
\BIBentrySTDinterwordspacing

\bibitem{Tonneau2018AnRobots}
S.~Tonneau, A.~Del~Prete, J.~Pettre, C.~Park, D.~Manocha, and N.~Mansard, ``{An Efficient Acyclic Contact Planner for Multiped Robots},'' \emph{IEEE Transactions on Robotics}, vol.~34, no.~3, pp. 586--601, 2018.

\bibitem{Kumagai2019BipedalFeasibility}
I.~Kumagai, M.~Morisawa, M.~Benallegue, and F.~Kanehiro, ``{Bipedal Locomotion Planning for a Humanoid Robot Supported by Arm Contacts Based on Geometrical Feasibility},'' in \emph{IEEE-RAS International Conference on Humanoid Robots}, vol. 2019-Octob.\hskip 1em plus 0.5em minus 0.4em\relax IEEE, 2019, pp. 132--139.

\bibitem{Kumagai2018EfficientMotion}
I.~Kumagai, M.~Morisawa, S.~Nakaoka, and F.~Kanehiro, ``{Efficient Locomotion Planning for a Humanoid Robot with Whole-Body Collision Avoidance Guided by Footsteps and Centroidal Sway Motion},'' in \emph{International Conference on Humanoid Robots}.\hskip 1em plus 0.5em minus 0.4em\relax IEEE, 2018, pp. 251--256.

\bibitem{Lin2020RobustPrediction}
Y.~C. Lin, L.~Righetti, and D.~Berenson, ``{Robust Humanoid Contact Planning with Learned Zero- A nd One-Step Capturability Prediction},'' \emph{IEEE Robotics and Automation Letters}, vol.~5, no.~2, pp. 2451--2458, 2020.

\bibitem{Ferrari2023Multi-contactFramework}
P.~Ferrari, L.~Rossini, F.~Ruscelli, A.~Laurenzi, G.~Oriolo, N.~G. Tsagarakis, and E.~M. Hoffman, ``{Multi-contact planning and control for humanoid robots: design and validation of a complete framework},'' \emph{Robotics and Autonomous Systems}, vol. 166, p. 104448, 2023.

\bibitem{Kingston2023ScalingSearch}
Z.~Kingston and L.~E. Kavraki, ``{Scaling Multimodal Planning: Using Experience and Informing Discrete Search},'' \emph{IEEE Transactions on Robotics}, vol.~39, no.~1, pp. 128--146, 2023.

\bibitem{Tonneau2018}
S.~Tonneau, P.~Fernbach, A.~Del~Prete, J.~Pettr{\'{e}}, and N.~Mansard, ``{2PAC: Two-point attractors for center of mass trajectories in multi-contact scenarios},'' \emph{ACM Transactions on Graphics}, vol.~37, no.~5, 2018.

\bibitem{Wang2020Multi-FidelityLocomotion}
J.~Wang, S.~Kim, S.~Vijayakumar, and S.~Tonneau, ``{Multi-Fidelity Receding Horizon Planning for Multi-Contact Locomotion},'' in \emph{IEEE-RAS International Conference on Humanoid Robots}.\hskip 1em plus 0.5em minus 0.4em\relax IEEE, 2020, pp. 53--60.

\bibitem{Winkler2018GaitParameterization}
A.~W. Winkler, C.~D. Bellicoso, M.~Hutter, and J.~Buchli, ``{Gait and Trajectory Optimization for Legged Systems Through Phase-Based End-Effector Parameterization},'' \emph{IEEE Robotics and Automation Letters}, vol.~3, no.~3, pp. 1560--1567, 2018.

\bibitem{Wang2022LearningPlanning}
J.~Wang, T.~S. Lembono, S.~Kim, S.~Calinon, S.~Vijayakumar, and S.~Tonneau, ``{Learning to Guide Online Multi-Contact Receding Horizon Planning},'' in \emph{IEEE International Conference on Intelligent Robots and Systems}.\hskip 1em plus 0.5em minus 0.4em\relax IEEE, 2022, pp. 12\,942--12\,949.

\bibitem{Sleiman2023VersatileLoco-Manipulation}
J.-P. Sleiman, F.~Farshidian, and M.~Hutter, ``{Versatile Multicontact Planning and Control for General Loco-Manipulation},'' \emph{Science Robotics}, vol.~8, no.~81, p. eadg5014, 2023.

\bibitem{Jelavic2023LSTP:Systems}
E.~Jelavic, K.~Qu, F.~Farshidian, and M.~Hutter, ``{LSTP: Long Short-Term Motion Planning for Legged and Legged-Wheeled Systems},'' \emph{IEEE Transactions on Robotics}, vol.~39, no.~6, pp. 4190--4210, 2023.

\bibitem{gcs}
T.~Marcucci, J.~Umenberger, P.~Parrilo, and R.~Tedrake, ``{Shortest Paths in Graphs of Convex Sets},'' \emph{SIAM Journal on Optimization}, vol.~34, no.~1, pp. 507--532, 2024.

\bibitem{McCrory2023GeneratingVisualization}
S.~McCrory, S.~Bertrand, A.~Mohan, D.~Calvert, J.~Pratt, and R.~Griffin, ``{Generating Humanoid Multi-Contact Through Feasibility Visualization},'' in \emph{IEEE-RAS International Conference on Humanoid Robots}.\hskip 1em plus 0.5em minus 0.4em\relax IEEE, 2023, pp. 1--8.

\bibitem{Wang2024OnlineApproximation}
J.~Wang, S.~Kim, T.~S. Lembono, W.~Du, J.~Shim, S.~Samadi, K.~Wang, V.~Ivan, S.~Calinon, S.~Vijayakumar, and S.~Tonneau, ``{Online Multi-Contact Receding Horizon Planning via Value Function Approximation},'' \emph{IEEE Transactions on Robotics}, vol.~40, pp. 2791--2810, 2024.

\bibitem{Zhao2022ReactiveEnvironments}
Y.~Zhao, Y.~Li, L.~Sentis, U.~Topcu, and J.~Liu, ``{Reactive task and motion planning for robust whole-body dynamic locomotion in constrained environments},'' \emph{International Journal of Robotics Research}, vol.~41, no.~8, pp. 812--847, 2022.

\bibitem{Ha2024Learning-basedPerspectives}
S.~Ha, J.~Lee, M.~Van De~Panne, Z.~Xie, W.~Yu, and M.~Khadiv, ``{Learning-based legged locomotion; state of the art and future perspectives},'' Tech. Rep., 2024.

\bibitem{Mastalli2020Crocoddyl:Control}
C.~Mastalli, R.~Budhiraja, W.~Merkt, G.~Saurel, B.~Hammoud, M.~Naveau, J.~Carpentier, L.~Righetti, S.~Vijayakumar, and N.~Mansard, ``{Crocoddyl: An Efficient and Versatile Framework for Multi-Contact Optimal Control},'' in \emph{Proceedings - IEEE International Conference on Robotics and Automation}, 2020, pp. 2536--2542.

\bibitem{brazil_omni3d}
G.~Brazil, A.~Kumar, J.~Straub, N.~Ravi, J.~Johnson, and G.~Gkioxari, ``{Omni3D: A Large Benchmark and Model for 3D Object Detection in the Wild},'' in \emph{Proceedings of the IEEE Computer Society Conference on Computer Vision and Pattern Recognition}, 2023, pp. 13\,154--13\,164.

\bibitem{Deits2015ComputingProgramming}
R.~Deits and R.~Tedrake, ``{Computing large convex regions of obstacle-free space through semidefinite programming},'' in \emph{Algorithmic Foundations of Robotics XI: Selected Contributions of the Eleventh International Workshop on the Algorithmic Foundations of Robotics}, vol. Springer I, 2015, pp. 109--124.

\bibitem{Radford2014Valkyrie:Robot}
N.~A. Radford, P.~Strawser, K.~Hambuchen, J.~S. Mehling, W.~K. Verdeyen, A.~S. Donnan, J.~Holley, J.~Sanchez, V.~Nguyen, L.~Bridgwater, and {Others}, ``{Valkyrie: NASA’s First Bipedal Humanoid Robot},'' \emph{Journal of Field Robotics}, vol.~32, no.~3, pp. 1--17, 2014.

\bibitem{Shah2023Towards2.0}
D.~Shah, M.~Savoldi, A.~Scalzo, A.~Mura, J.~Losi, V.~Gaggero, L.~Fiorio, and M.~Maggiali, ``{Towards design and development of new joint modules for humanoid ergoCub 2.0},'' in \emph{IEEE-RAS International Conference on Humanoid Robots}.\hskip 1em plus 0.5em minus 0.4em\relax IEEE, 2023, pp. 1--7.

\bibitem{Yoshikawa1985ManipulabilityMechanisms.}
T.~Yoshikawa, ``{Manipulability of Robotic Mechanisms.}'' \emph{The International Journal of Robotics Research}, vol.~4, no.~2, pp. 3--9, 1985.

\bibitem{Carpentier2018MulticontactRobots}
J.~Carpentier and N.~Mansard, ``{Multicontact Locomotion of Legged Robots},'' \emph{IEEE Transactions on Robotics}, vol.~34, no.~6, pp. 1441--1460, 2018.

\bibitem{WernerICRA24Cliques}
P.~Werner, A.~Amice, T.~Marcucci, D.~Rus, and R.~Tedrake, ``{Approximating Robot Configuration Spaces with few Convex Sets using Clique Covers of Visibility Graphs},'' in \emph{IEEE International Conference on Robotics and Automation}, 2024.

\end{thebibliography}

\end{document}